\documentclass[10pt,twocolumn,letterpaper]{article}

\usepackage{multicol}
\usepackage{multirow}

\usepackage{algorithmic}

\usepackage[pagenumbers]{iccv} 

\usepackage[dvipsnames]{xcolor}
\usepackage[accsupp]{axessibility}

\makeatletter
\DeclareRobustCommand\onedot{\futurelet\@let@token\@onedot}
\def\@onedot{\ifx\@let@token.\else.\null\fi\xspace}

\makeatother

\newcommand{\boldparagraph}[1]{\vspace{0.3em}\noindent{\bf #1.}}

\renewcommand{\paragraph}[1]{\boldparagraph{#1}}

\definecolor{darkgreen}{rgb}{0,0.7,0}
\definecolor{newyellow}{rgb}{1,0.8,0.05}
\definecolor{newgreen}{rgb}{0.2,0.8,0.2}
\definecolor{Gray}{gray}{0.85}

\definecolor{myorange}{RGB}{255, 127, 102}
\definecolor{mygreen}{RGB}{120, 210, 170}
\definecolor{myblue}{RGB}{0, 102, 204}
\definecolor{lightblue}{RGB}{232, 244, 248}
\definecolor{myred}{RGB}{220, 60, 60}
\definecolor{mypurple}{RGB}{160, 130, 210}
\definecolor{mypink}{RGB}{220, 100, 120}

\makeatletter
\def\adl@drawiv#1#2#3{%
        \hskip.5\tabcolsep
        \xleaders#3{#2.5\@tempdimb #1{1}#2.5\@tempdimb}%
                #2\z@ plus1fil minus1fil\relax
        \hskip.5\tabcolsep}
\newcommand{\cdashlinelr}[1]{%
  \noalign{\vskip\aboverulesep
           \global\let\@dashdrawstore\adl@draw
           \global\let\adl@draw\adl@drawiv}
  \cdashline{#1}
  \noalign{\global\let\adl@draw\@dashdrawstore
           \vskip\belowrulesep}}
\makeatother

\definecolor{iccvblue}{rgb}{0.21,0.49,0.74}
\usepackage[pagebackref,breaklinks,colorlinks,allcolors=iccvblue]{hyperref}

\def\paperID{11762} 
\def\confName{ICCV}
\def\confYear{2025}

\title{Curve-Aware Gaussian Splatting for 3D Parametric Curve Reconstruction}

\author{
Zhirui Gao$^{*}$  \quad 
Renjiao Yi$^{*}$  \quad
Yaqiao Dai  \quad
Xuening Zhu \quad
 Wei Chen  \quad
Chenyang Zhu$^{\dagger}$  \quad
Kai Xu$^{\dagger}$ \\
 National University of Defense Technology\\
{\href{https://zhirui-gao.github.io/CurveGaussian/}{zhirui-gao.github.io/CurveGaussian}}
}

\begin{document}
\maketitle

\let\thefootnote\relax\footnotetext{$^*$Co-first authors \quad $^\dag$Corresponding authors}

\begin{abstract}
This paper presents an end-to-end framework for reconstructing 3D parametric curves directly from multi-view edge maps. Contrasting with existing two-stage methods that follow a sequential ``edge point cloud reconstruction and parametric curve fitting'' pipeline, our one-stage approach optimizes 3D parametric curves directly from 2D edge maps, eliminating error accumulation caused by the inherent optimization gap between disconnected stages. 
However, parametric curves inherently lack suitability for rendering-based multi-view optimization, necessitating a complementary representation that preserves their geometric properties while enabling differentiable rendering. We propose a novel bi-directional coupling mechanism between parametric curves and edge-oriented Gaussian components. This tight correspondence formulates a curve-aware Gaussian representation, \textbf{CurveGaussian}, that enables differentiable rendering of 3D curves, allowing direct optimization guided by multi-view evidence. Furthermore, we introduce a dynamically adaptive topology optimization framework during training to refine curve structures through linearization, merging, splitting, and pruning operations. Comprehensive evaluations on the ABC dataset and real-world benchmarks demonstrate our one-stage method's superiority over two-stage alternatives, particularly in producing cleaner and more robust reconstructions. Additionally, by directly optimizing parametric curves, our method significantly reduces the parameter count during training, achieving both higher efficiency and superior performance compared to existing approaches.
\end{abstract}

\section{Introduction}
\label{sec:intro}

\begin{figure}[tbh]
  \centering
  \includegraphics[width=\linewidth]{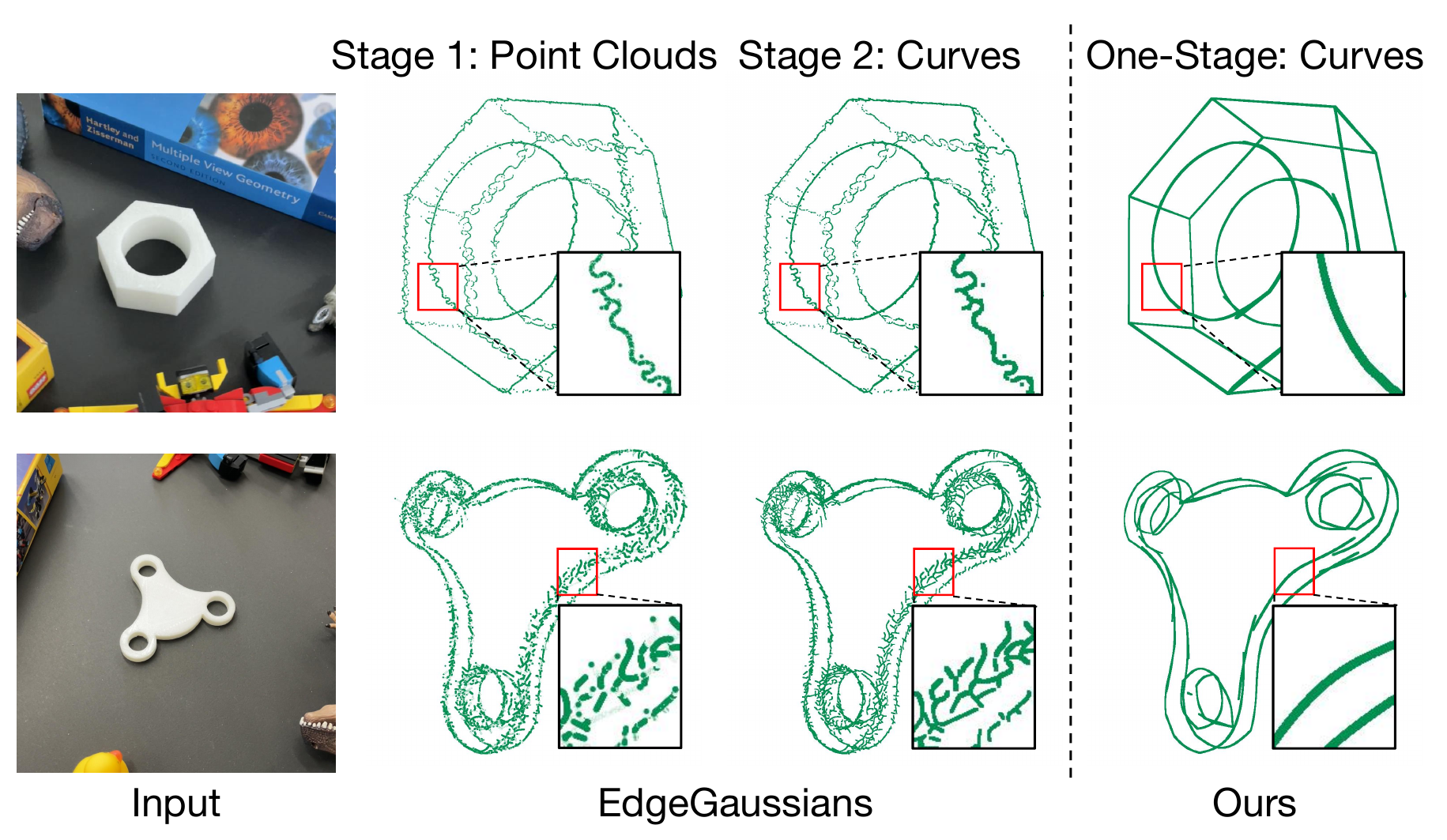}
  \vspace{-12px}
  \caption{Our method achieves clean and accurate parametric curve reconstruction through a one-stage optimization utilizing the proposed curve-aware Gaussians. It eliminates the limitations of existing two-stage approaches, 
  where the accuracy of the stage-2 curve fitting is heavily dependent on the quality of the initial point cloud reconstruction in stage-1. 
  }
  \label{fig:teaser}
  \vspace{-12px}
\end{figure}

\begin{figure*}[tbh]
  \centering
  \includegraphics[width=1.0\textwidth]{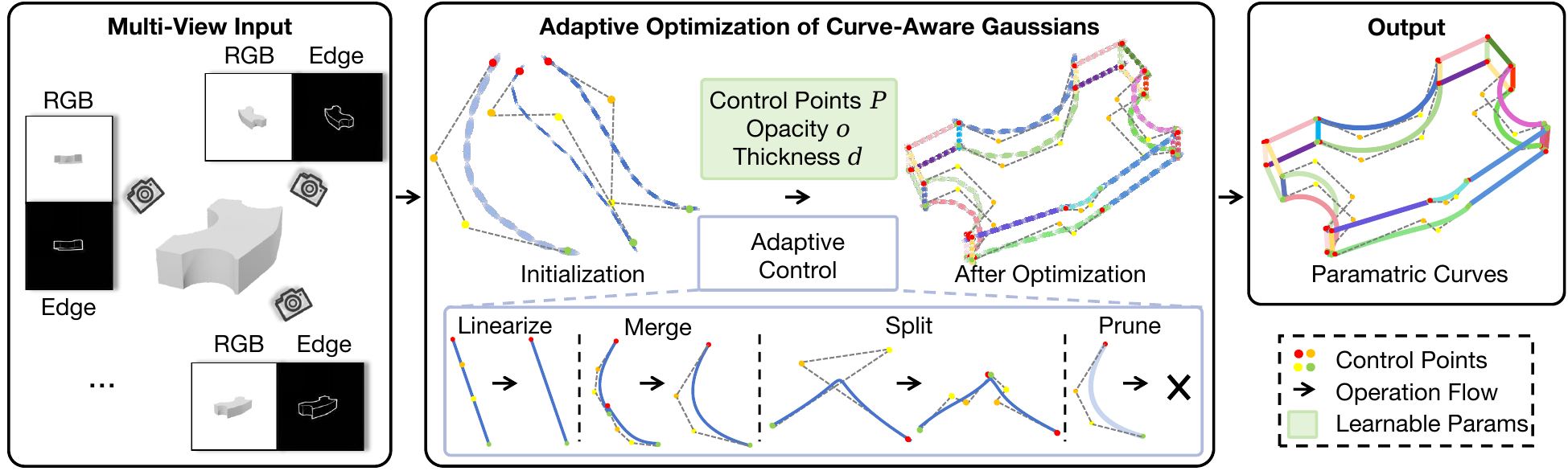}
  \caption{\textbf{Overview of CurveGaussian.} We propose a \textbf{curve-aware Gaussian} representation that optimizes parametric curves through a one-stage optimization in a self-supervised manner by re-rendering losses. The method employs many adaptive strategies, including curve linearization, merging, splitting, and pruning, to dynamically adjust the curves during training. }
  \label{fig:ppl}
  \vspace{-6px}
\end{figure*}

Parametric curves, as fundamental geometric primitives, play a vital role in structured 3D representations and geometric understanding~\cite{liu2023limap,wang2025omniearth,gao2024learning,lan2022arm3d}. Their ability to compress geometric information while maintaining CAD compatibility~\cite{Wang_2024} makes them indispensable for industrial applications. A critical challenge lies in reconstructing high-precision parametric curves from unstructured data, such as noisy point clouds or multi-view images, as sensor imperfections and incomplete observations~\cite{wang2025harnessingmassivesatelliteimagery,10506102} inevitably degrade reconstruction quality.

Early methods~\cite{liu2021pc2wf,wang2020pie,zhu2023nerve} for 3D curve reconstruction typically require clean 3D point clouds as input for subsequent fitting procedures. The recent NEF framework~\cite{Ye_2023_CVPR} first achieves 3D curve reconstruction directly from 2D images through a two-stage ``edge point cloud reconstruction + parametric curve fitting'' pipeline that combines neural radiance fields (NeRFs)~\cite{mildenhall2020nerf} with RANSAC-based optimization~\cite{Fischler1981RandomSC}. However, this approach accumulates errors from the disconnected two-stage optimization process, making it sensitive to the noise of the reconstructed point cloud. EMAP~\cite{Li2024CVPR} improves upon NEF by enhancing edge localization through the unsigned distance function (UDF) modeling and unbiased equation rendering. Though its ``cluster-then-fitting'' strategy reduces error propagation, the method remains fundamentally constrained by the point cloud quality given in the first stage. Notably, these approaches suffer from low computational efficiency due to NeRF optimizations~\cite{mildenhall2020nerf}. While EdgeGaussians~\cite{chelani2024edgegaussians} accelerates reconstruction using Gaussian splatting~\cite{kerbl3Dgaussians}, it still fails to overcome the inherent limitations of a two-stage fashion pipeline.

The two-stage paradigm introduces two fundamental limitations. First, the parametric fitting process is strongly coupled with observation noises/errors, propagated from 2D edge detection to 3D point clouds through multi-view triangulation. It results in redundant branches or breaks in final curves (\textbf{as shown in Fig.~\ref{fig:teaser}}). Secondly, greedy iterative fitting (e.g., RANSAC~\cite{Fischler1981RandomSC}) produces local optima accumulation, causing curve redundancy to grow exponentially with scene complexity, which severely impacts downstream application efficiency. Eliminating the optimization gap between stages emerges as the key to robust reconstruction.

While 2D edge maps are much easier to acquire, bypassing the explicit 3D edge point cloud reconstruction is a promising way to solve this problem. Direct one-stage curve optimization confronts two fundamental challenges: parametric representations inherently lack differentiable rendering capability, while neural rendering frameworks (NeRF/Gaussians) cannot preserve geometric continuity when approximating curves. We present a hybrid neural representation, \textbf{CurveGaussian}, that combines parametric curves with edge-oriented Gaussian components. By establishing bi-directional coupling, each Bézier curve segment precisely controls the spatial distribution of its associated Gaussian components. It enables differentiable rendering through splatting while significantly reducing the parameter count required for the optimization. 

Building upon this hybrid representation, this paper proposes a one-stage method for parametric curve reconstruction from multi-view edge maps. The core concept bypasses intermediate point cloud representations to directly perform differentiable optimization on the control points of Bézier curves, where each curve comprises multiple edge-oriented Gaussians. Specifically, each Gaussian component is bound to a specific edge segment of its corresponding curve, with positional and orientational attributes analytically derived from curve parameters. Rendering losses are backpropagated to update curve control points, enabling simultaneous exploitation of neural rendering's photometric consistency constraints and parametric curves' intrinsic geometric regularization. To further optimize curve topology during training for cleaner and more compact reconstructions, we propose a dynamic topology optimization strategy that adaptively refines edge structures through merging, splitting, and pruning operations. Additionally, we improve the optimization loss to prioritize the structural compactness of generated curves. Experiments demonstrate that our one-stage method increases 14.5\% in reconstruction accuracy, improves computational efficiency by 33\%, and reduces generated curve counts by 70.5\% compared with two-stage SOTA alternatives, attaining more accurate and compact reconstructions. In summary, the contributions include: 
\begin{itemize}
    \item We propose a novel bi-directional coupling framework between parametric curves and edge-oriented Gaussian components, enabling direct optimization of parametric curves through differentiable Gaussian splatting. 
    \item We propose a training-time structural refinement strategy that adaptively optimizes curve topology, achieving compact parametric curve reconstructions. 
    \item Our one-stage optimization framework demonstrates superior robustness against reconstruction noise compared to existing two-stage methods, delivering significantly improved accuracy in real-world scenarios.

\end{itemize}

\section{Related Works}

\subsection{3D Parametric Curve Reconstruction}
Early methods~\cite{sinha2005multi, schindler2006line, chandraker2009moving} reconstruct 3D line segments using Structure-from-Motion (SfM), involving line detection~\cite{pautrat2021sold2, pautrat2023deeplsd, xue2023holistically}, matching~\cite{abdellali2021l2d2, pautrat2023gluestick}, and triangulation~\cite{bartoli2005structure, baillard1999automatic, chandraker2009moving, liu2023limap}. 
Other works address line mapping by integrating graph clustering~\cite{hofer2017efficient} and planar constraints~\cite{wei2022elsr} into SLAM systems~\cite{lim2022uv,marzorati2007integration, shu2023structure}. NEAT~\cite{xue2024neat} optimizes 3D line junctions using VolSDF~\cite{yariv2021volume} but is limited to straight lines and textured objects. Recent advances in 3D parametric curve reconstruction focus on two paradigms: point cloud-based and image-based approaches. Point cloud-based techniques~\cite{liu2021pc2wf,wang2020pie,zhu2023nerve,matveev2022def} extract parametric curves directly from edge point clouds, but their performance tends to degrade with noisy input. Alternatively, image-based methods have emerged as a promising solution, leveraging powerful radiance fields such as NeRF~\cite{mildenhall2020nerf} and 3DGS~\cite{kerbl3Dgaussians}. These approaches benefit from the remarkable success of neural representations across various domains~\cite{gao2025generic,gao2024fdc,zhou2025monomobilityzeroshot3dmobility,gao2025selfsupervisedlearninghybridpartaware}, enabling high-quality 3D edge reconstruction from multi-view images that are typically pre-processed using edge detection~\cite{canny1983variational,su2021pixel,10168686}. Among these, NEF~\cite{Ye_2023_CVPR} and EMAP~\cite{Li2024CVPR} reconstruct 3D edge fields from 2D images using the implicit radiance field, using clustering~\cite{Fischler1981RandomSC,ICL-SSL, DealMVC} to extract parametric edges.  More recently, EdgeGaussians~\cite{chelani2024edgegaussians} leverages 3DGS for efficient edge field reconstruction.  Despite these advances, these works are constrained by a two-stage optimization paradigm, where the second stage of parametric edge extraction is prone to inaccuracies in edge representation and often produces fragmented and redundant edges. To address these issues, this paper proposes a one-stage parametric curve optimization pipeline, enabling more accurate and compact parametric edge extraction.

\subsection{Curve-based Differentiable Rendering }

Previous works have explored differentiable rendering of curves. One representative work is DiffVG~\cite{Li:2020:DVG}, which proposes a differentiable 2D vector graphics rasterizer that optimizes vectorized images using 2D Bézier curves and other parametric shapes. DRPG~\cite{worchel:2023:drpg} proposes an efficient approach for differentiable rendering of 3D parametric surfaces and curves, achieved by approximating continuous parametric representations using a triangle mesh. Recently, 3Doodle~\cite{3Doodle} and Diff3DS~\cite{zhang2024diff3ds} project 3D Bézier curves onto 2D Bézier curves, which are then rendered using the differentiable rasterizer from DiffVG. They achieve significant progress in 3D sketch generation by leveraging differentiable curve rendering. In contrast, rather than projecting Bézier curves onto 2D planes, we directly project the coupled Gaussians from the curves. By leveraging the high-quality rendering capabilities of Gaussians and backpropagating gradients, we propose to direct the optimization of parametric curves, which is demonstrated to be effective, efficient, and straightforward to implement.

\section{Method}
Most existing methods~\cite{Ye_2023_CVPR, Li2024CVPR, chelani2024edgegaussians} for reconstructing parametric 3D curves from multi-view edge images usually follow a two-stage approach: reconstructing a 3D edge representation (e.g., using NeRF or 3DGS) to get edge point clouds, then fitting parametric curves. 
While this approach is conceptually simple, it suffers from critical limitations. 
The two-stage paradigm leads to error accumulations. Since curve fitting heavily depends on the accuracy of the point cloud reconstruction, any noise from the first stage significantly degrades the final parametric curves. These issues also lead to lower efficiency. 
To address these limitations, we propose a one-stage optimization approach that directly optimizes parametric curves by the proposed curve-aware Gaussian splatting as illustrated in Fig.~\ref{fig:ppl}. 

We begin by formalizing the problem in Section~\ref{subsec:definition}, followed by introducing the curve-aware Gaussian splatting framework in Section~\ref{subsec:curveGS}, which enables direct rendering of parametric curves.  The adaptive optimization is then presented in Section~\ref{subsec:adaptive} and Section~\ref{subsec:optimize}, aimed at stable learning of complex curve topologies during training.

\subsection{Problem Definition}
\label{subsec:definition}

The task of \textbf{parametric curve reconstruction} aims to recover 3D parametric curves from multi-view 2D edge maps. Formally, given a set of parametric curves $\{\mathcal{C}_j\}$, where $\mathcal{C}_j$ denotes the $j$-th curve, we optimize their parameters (e.g., control points) such that their rendered 2D projections align with the observed edge images across all views. It is formulated as the following optimization:
\begin{equation}
\label{equ:definition}
\arg\min_{\{\mathcal{C}_j\}} \sum_{k=0}^{K-1} \sum_{j=0}^{C-1} \mathcal{L}_r\left(f(\{\mathcal{C}_j\}, \mathbf{T}_k), \hat{I}_k\right).
\end{equation}
Here, $\mathbf{T}_k$ denotes the camera pose for the $k$-th view, $\hat{I}_k$ is the ground-truth edge map for view $k$, $f(\cdot)$ is the rendering function that maps parametric curves to edge maps, and $\mathcal{L}_r$ is the rendering loss. $K$ and $C$ represent the number of observed views and the total number of curves, respectively.

By leveraging a differentiable rendering function $f(\cdot)$, our approach tightly couples curve reconstruction with multi-view edge images, and 
directly optimizes parametric curves, enforcing local edge consistency. It avoids the error accumulation and inefficiencies in two-stage processes. Additionally, directly optimizing curve parameters reduces the number of optimization variables and improves computational efficiency, as demonstrated in Tab.~\ref{tab:ABC_2}. 

In the framework, we represent each curve $\mathcal{C}_j$ as a parametric curve $\mathbf{c}_j(t)$, where $t \in [0,1]$ parameterizes the curve from start point ($t=0$) to end point ($t=1$). We use two types of parametric curves: cubic and first-order Bézier curves, which are sufficient to describe most shapes encountered in most scenarios. 

\textbf{Cubic Bézier Curves:}
A cubic Bézier curve is defined by four control points: $\mathbf{P}_j^0$ (start point), $\mathbf{P}_j^1$, $\mathbf{P}_j^2$, and $\mathbf{P}_j^3$ (end point). The curve is parameterized as:
\begin{equation}
\mathbf{c}_j(t) = \sum_{k=0}^3 B_k^3(t)\mathbf{P}_j^k,
\end{equation}
where $B_k^3(t) = \binom{3}{k}t^k(1-t)^{3-k}$ are the Bernstein basis functions. The control points determine the shape of the curve, allowing for smooth and flexible representations.

\textbf{First-order Bézier Curves (Line Segments):}
A First-order Bézier curve (line segment) is defined by its two endpoints, $\mathbf{P}_j^0$ and $\mathbf{P}_j^1$, and is parameterized as:
\begin{equation}
\mathbf{c}_j(t) = (1-t)\mathbf{P}_j^0 + t\mathbf{P}_j^1.
\end{equation}
First-order Bézier curves are simpler than cubic Bézier curves but are useful for representing straight sections. 

To better represent curves in 3D space, ensure compatibility with Gaussian representations, and enhance the expressiveness of curves, each curve is associated with two additional parameters: opacity $o_j \in [0,1]$, which controls the visibility of the $j$-th curve, and thickness $d_j \in \mathbb{R}^+$, which determines the width of the curve.

\subsection{Curve-Aware Gaussian Splatting}
\label{subsec:curveGS}
The key challenge lies in designing a differentiable rendering function $f(\cdot)$. To achieve this, we propose a hybrid representation combining parametric curves with Gaussian primitives. We establish a bi-directional coupling between parametric curves and 3D Gaussians, where each curve $c_j$ generates $N$ ($N$=12) Gaussians through uniform sampling as in Fig.~\ref{fig:ppl}. Gaussian attributes are determined by curve geometry: 
\begin{itemize}
    \item {\textbf{Position \& orientation}}: The $i$-th Gaussian $\mathcal{G}_j^i$ in $c_j$ is anchored at the curve point $\mathbf{p}_j(t_i)$, where $t_i = \frac{i+0.5}{N}$ for $i \in \{0, 1, \ldots, N-1\}$. The principal axis $\mathbf{v}_0$ of $\mathbf{p}_j(t_i)$ is aligned with the curve tangent $\mathbf{T}_j(t_i)$. The second axis $\mathbf{v}_1$ is adjusted to be perpendicular to $\mathbf{v}_0$, and the third axis $\mathbf{v}_2$ is obtained as the cross product of $\mathbf{v}_0$ and $\mathbf{v}_1$, ensuring orthogonality in 3D space.

\item {\textbf{Scales \& opacity}}: The axial scales are defined as: $\mathbf{s}^{j,i} = \left[\|\Delta\mathbf{p}_j^{i}\|, d_j, d_j\right]^\top$, where $\Delta\mathbf{p}_j^{i} = \mathbf{p}_j(t_{i+1}) - \mathbf{p}_j(t_i)$, and $d_j$ denotes a curve-specific thickness parameter. In practice, the scale of the principal axis is significantly larger than the other axes, resulting in rod-shaped and edge-orientated Gaussians. The opacity $o_j^i$ of the Gaussians is inherited from the parent curve's attribute $o_j$.

\item {\textbf{Mask}}: To address redundancy in overlapping regions and potential over-coverage in non-edge areas, we introduce a learnable importance mask $m^{j,i} \in [0,1]$ for each Gaussian following \cite{lee2024c3dgs}. The mask operates as a gating mechanism during optimization, which means
$\mathcal{G}_j^i$ is effectively suppressed when $m^{j,i}$ is close to 0. This attribute automatically indicates redundant curve segments by gradient-driven mask decay.

\end{itemize}

As a result, the rendering function $f(\cdot)$ in Equ.~\ref{equ:definition} can be implemented by curve-aware Gaussian splatting. The splatting follows the standard 3DGS pipeline~\cite{kerbl3Dgaussians} but removes the color attribute, as the edge map is single-channel. Instead, alpha-weighted values are directly used to compute the rendered image and the rendering loss.

\begin{figure}[t]
  \centering
  \includegraphics[width=\linewidth]{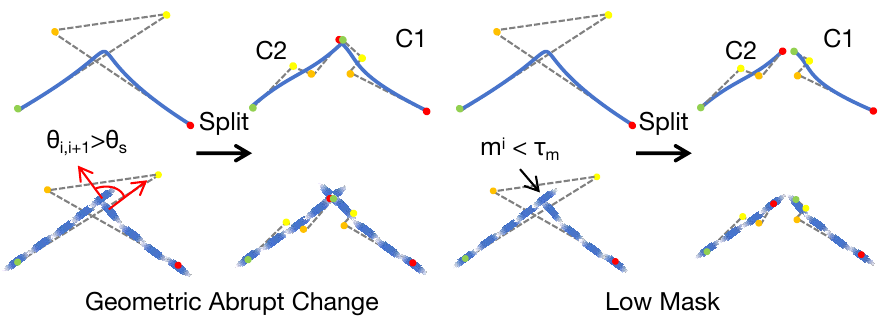}
  \caption{Visualization of splitting strategy. The first row demonstrates on parametric curves, and the second row demonstrates corresponding strategies on the proposed curve-aware Gaussians.}
  \label{fig:control}
  \vspace{-10px}
\end{figure}

\subsection{Adaptive Control of Curve-Aware Gaussians}
\label{subsec:adaptive}
While lacking prior knowledge of the number or locations of curves, we initialize the process with random Bézier parameterized curves. This approach leads to an optimization process that differs significantly from the original 3DGS. Specifically, the number of Gaussian primitives evolves dynamically, starting from a large set and gradually reducing to a smaller, more refined set during optimization. This behavior necessitates an adaptive strategy to ensure stable and accurate results. Four adaptive strategies are employed, as illustrated in the bottom part of Fig.~\ref{fig:ppl}, with details below:

\paragraph{Curve Linearizing Strategy} 
During training, cubic Bézier curves are converted into first-order Bézier curves (line segments) if they are close to straight. 
Specifically, by uniformly sampling the cubic Bézier curve $\mathbf{c}_j(t)$, we obtain a set of points $\{\mathbf{p}_i \}_{i=1}^N$, and then fit a straight line. The curve $\mathbf{c}_j(t)$ is replaced by a line segment $\mathbf{c}'_j(t)$ if satisfying :

\vspace{-10px}
\begin{equation}
\begin{split}
  \mathbf{c}'_j(t) = (1-t)\mathbf{P}_j^0 + t\mathbf{P}_j^3, \\ \quad \text{if } \frac{1}{N} \sum_{i=1}^N \| \mathbf{c}'_j(t_i) - \mathbf{p}_i \|_2 < \tau_l,
\end{split}
\end{equation}
\vspace{-10px}

\noindent where $\tau_l$ is a predefined error threshold, $\mathbf{P}_j^0$ and $\mathbf{P}_j^3$ are the start and end points of the cubic Bézier curve $\mathbf{c}_j(t)$, and $t \in [0, 1]$. This transformation reduces computational complexity while preserving the precise geometry of the structure.

\paragraph{Curve Merging Strategy}
\begin{itemize}
    \item \textbf{Line Merging}: For two line segments $\mathbf{c}_1$ and $\mathbf{c}_2$, the angle $\theta$ between their directions and the spatial distance $d$ between their closest endpoints are computed. The lines are merged if $\theta < \tau_{la} \text{ and } d < \tau_{ld}$, where $\tau_{la}$ and $\tau_{ld}$ are predefined angular and spatial thresholds.

    \item \textbf{Cubic Bézier Curve Merging}: For two adjacent Bézier curves $ \mathbf{c}_1 $ and $ \mathbf{c}_2 $, a new Bézier curve $ \mathbf{c}' $ is fitted through a set of sampled points $ \{ \mathbf{p}_i \}_{i=1}^N $ from both curves. The two curves are merged and replaced with $ \mathbf{c}' $ if $ \frac{1}{N} \sum_{i=1}^N \| \mathbf{c}'(t_i) - \mathbf{p}_i \|_2 < \tau_{b}$, where $ \tau_{b} $ is a predefined error threshold.
\end{itemize}
This strategy not only ensures the curves are compact but also maintains geometric accuracy, making it efficient for curve-aware Gaussian splatting.

\paragraph{Curve Splitting Strategy} 
A cubic Bézier curve $\mathbf{c}_j$, with its control points $\mathbf{P}_{j,0}^0, \mathbf{P}_{j,0}^1, \mathbf{P}_{j,0}^2, \mathbf{P}_{j,0}^3$, can be split at $t=s$ into two sub-curves using de Casteljau's algorithm:
        \begin{equation}
            \begin{aligned}
                c_1: & \{\mathbf{P}_{j,0}^0, \mathbf{P}_{j,1}^0, \mathbf{P}_{j,2}^0, \mathbf{P}_{j,3}^0\} \\
                c_2: & \{\mathbf{P}_{j,3}^0, \mathbf{P}_{j,2}^1, \mathbf{P}_{j,1}^2, \mathbf{P}_{j,0}^3\} ,
            \end{aligned}
        \end{equation}
        the intermediate control points are computed recursively:
        \begin{equation}
            \mathbf{P}_{j,k}^i = (1-s)\mathbf{P}_{j,k-1}^i + s\mathbf{P}_{j,k-1}^{i+1}.
        \end{equation}
As in Fig.~\ref{fig:control}, splitting operation is triggered in two cases:
\begin{itemize}
    \item \textbf{Geometric abrupt change detection}: Local geometric abrupt changes are detected by computing the angle $\theta_{i,i+1} = \arccos(\mathbf{v}_1^{j,i} \cdot \mathbf{v}_1^{j,i+1})$ between the principal axes of adjacent Gaussians on a curve. If $\max_i\theta_{i,i+1} > \theta_{\text{s}}$, locate the point of maximum abrupt change as $t^* = \arg\max_t\theta(t)$ and split the curve directly at $t^*$ to obtain two new curves $c_1$ and $c_2$.
    \item \textbf{Low mask detection}: If the Gaussian with the smallest mask value on a curve satisfies $m_j^i < \tau_{m}$, indicating insufficient occupy, the curve is split at $t_{i-1}$ to retain $c_1$ and at $t_{i+1}$ to retain $c_2$, with the unreliable segment centered at $t_i$ removed.

\end{itemize}
This strategy ensures that the curve retains its smoothness while eliminating redundant segments.

\paragraph{Curve Pruning Strategy} 
A curve $c_j$ is dynamically removed if $o_j < \tau_d$,
where $o_j$ denotes the opacity of the $j$-th curve, and $\tau_d$ is a predefined threshold. If the masks of all Gaussians on the curve are smaller than $\tau_{m}$, the curve will also be removed. This strategy ensures the removal of curves that are essentially transparent, thereby improving the overall quality of the representation.

\subsection{Optimizing Curve-Aware Gaussians}
\label{subsec:optimize}

We optimize curve parameters $\{\mathbf{P}_j^k, o_j, d_j\}$ through the following multi-objective loss:

\paragraph{Edge-Aware Rendering Loss} The original loss function in 3DGS for training 3D Gaussians is defined as:
\begin{equation}
\mathcal{L}_{I} = (1 - \lambda) \mathcal{L}_1 + \lambda \mathcal{L}_{D\text{-}SSIM},
\end{equation}
where $\mathcal{L}_1$ and $\mathcal{L}_{D\text{-}SSIM}$ are supervised by the ground-truth image. However, this formulation suffers from gradient collapse in edge map optimization due to the sparse distribution of edge pixels, often leading to the transparency of all Gaussians converging to zero.

To address this issue, we propose a novel edge-aware loss~\cite{Ye_2023_CVPR} function $\mathcal{L}_{edge}$ that replaces $\mathcal{L}_1$ and balances the influence of edge and non-edge pixels in edge maps. The loss function is defined as:

\begin{equation}
\mathcal{L}_{edge} = \frac{|M_I|}{|E_I|} \sum_{i \in N_I} \|I_i - \hat{I}_i\|_2^2 +  \frac{|N_I|}{|E_I|} \sum_{i \in M_I} \|I_i - \hat{I}_i\|_2^2.
\end{equation}

\begin{itemize}
    \item $M_I$ denotes the set of edge pixels in the current edge map (pixels with intensity greater than the threshold $\eta = 0.1$), and vice versa for $N_I$, and $E_I$ for all pixels.
    \item $I_i$ and $\hat{I}_i$ are the pixel values in the rendered image and the ground-truth edge map, respectively.
\end{itemize}
This function enhances the model's ability to discriminate edge pixels by adaptively balancing the contributions of edge and non-edge regions during optimization.

\paragraph{Smooth Connection Regularization}  
To ensure smooth connections between learned curves, we design a regularization term that minimizes the distance between adjacent endpoints. Let $P_i$ and $P_j$ denote the endpoints (either start or end) of the $i$-th and $j$-th curves, respectively. The endpoint connection regularization term is defined as:
\begin{equation}
\mathcal{L}_{conn} = \sum_{i=1}^{C} \sum_{j=1}^{C} \mathbb{I}(\| P_i - P_j \|_2 < \tau) \cdot \| P_i - P_j \|_2^2,
\end{equation}
where $C$ is the total number of curves, $\tau$ is a distance threshold, and $\mathbb{I}(\cdot)$ is the indicator function that equals 1 if $\| P_i - P_j \|_2 < \tau$ and 0 otherwise.  

\paragraph{Smooth Curve Regularization}
To ensure smoother curve trajectories, we introduce an additional loss term as the curve smoothness constraint. This constraint is achieved by minimizing the difference between Gaussian direction vectors of adjacent sample points. It is defined as:
\begin{equation}
\mathcal{L}_{smo} = \sum_{j=1}^{C} \sum_{i=1}^{N_j-1} \| \mathbf{v}_1^{j,i} - \mathbf{v}_1^{j,i+1} \|_2^2,
\end{equation}
where $N_j$ is the number of sample points on the $j$-th curve, $\mathbf{v}_1^{j,i}$ denotes the Gaussian direction vector at the $i$-th sample point of the $j$-th curve.

\begin{figure*}[tbh]
  \centering
  \includegraphics[width=0.96\linewidth]{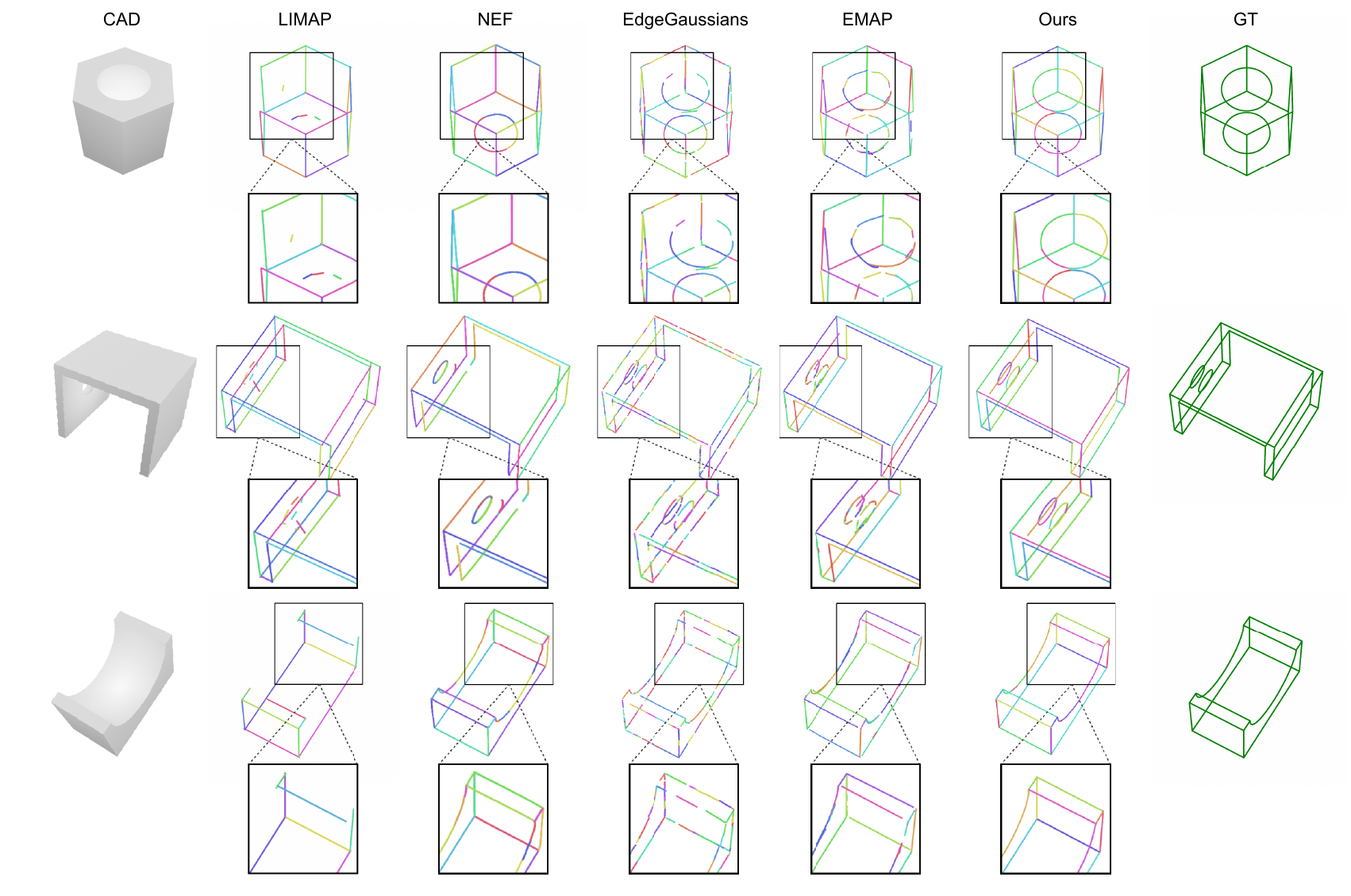}
  \vspace{-1.3em}
  \caption{\textbf{Qualitative comparisons on ABC-NEF~\cite{Ye_2023_CVPR}.} Distinct colors represent different curves/lines. Our method achieves more complete and accurate edge reconstruction of objects while maintaining parametric compactness.}
  \label{fig:abc_vis}
  \vspace{-10px}
\end{figure*}

\paragraph{Concise Curve Regularization}
To promote concise curve representations and enhance optimization convergence, we introduce an additional regularization term $\mathcal{L}_{reg}$ to constrain the opacity of the curves:
\begin{equation}
\mathcal{L}_{reg} = \sum_{j=1}^{C} \log\left(1 + \frac{o_j^2}{0.5}\right),
\end{equation}
where $o_j$ denotes the opacity of the $j$-th curve.
Furthermore, to eliminate redundant Gaussian curves, we adopt a masking loss $\mathcal{L}_{m}$, inspired by ~\cite{lee2024c3dgs}.


In summary, the final loss function for training the parametric curves is defined as follows:
\begin{equation}
\label{equ:loss}
\mathcal{L}_{all} =  \mathcal{L}_{edge} + \lambda _1\mathcal{L}_{conn} + \lambda _2  \mathcal{L}_{smo} + \lambda_3 \mathcal{L}_{reg} + \lambda_4 \mathcal{L}_{m}.
\end{equation}
$\mathcal{L}_r$ in Equ.~\ref{equ:definition} is implemented as $\mathcal{L}_{all}$ here.

\section{Experiments}

\subsection{Experiment Settings}

\textbf{Datasets.} 
We evaluate the method on three public benchmarks: ABC-NEF~\cite{Ye_2023_CVPR},  MV2Cyl's real objects~\cite{hong2024mv2cyl}, and Replica~\cite{replica19arxiv}.  
Consistent with EMAP~\cite{Li2024CVPR} and EdgeGaussians~\cite{chelani2024edgegaussians}, we utilize a selected subset of the ABC-NEF dataset, comprising 82 ABC~\cite{koch2019abc} models. Each model provides parametric ground-truth edges for quantitative 3D edge evaluation, along with 50 rendered views that capture diverse object perspectives. MV2Cyl~\cite{hong2024mv2cyl} introduces a collection of real-world objects created by 3D-printing various sketch-extrude models from Fusion360~\cite{willis2021fusion}, providing multi-view images, camera poses, and CAD models. We demonstrate 6 objects to evaluate the performance of real-world objects. For Replica, each scene offers 100 multi-view edge maps for a thorough evaluation. Given the lack of ground-truth edge annotations, the analysis on Replica focuses primarily on qualitative comparisons to assess the effectiveness for complex scenes.

\noindent \textbf{Baselines.} The method is compared with four state-of-the-art baselines for 3D line and curve reconstruction, which include three learning-based methods, NEF~\cite{Ye_2023_CVPR}, EdgeGaussians~\cite{chelani2024edgegaussians}, and EMAP~\cite{Li2024CVPR}, and one line-based Structure-from-Motion (SfM) method LIMAP~\cite{liu2023limap}.

\begin{table*}[!t]
\centering
\scalebox{0.85}{
\begin{tabular}{c|c|c|cc|ccc|ccc|ccc}
     \hline 
  Method &
  Detector &
  Modal &
  Acc.$\downarrow$ &
  Comp.$\downarrow$ &
  $\text{R}_5 \uparrow$ &
  $\text{R}_{10}\uparrow$ &
  $\text{R}_{20} \uparrow$ &
  $\text{P}_5 \uparrow$ &
  $\text{P}_{10} \uparrow$ &
  $\text{P}_{20} \uparrow$ &
  $\text{F}_5 \uparrow$ &
  $\text{F}_{10} \uparrow$ &
  $\text{F}_{20}\uparrow$ \\ \hline
  \multirow{2}{*}{LIMAP~\cite{liu2023limap}} &
  LSD &
  Line &
  9.9 &
  18.7 &
  36.2 &
  82.3 &
  87.9 &
  43.0 &
  87.6 &
  93.9 &
  39.0 &
  84.3 &
  90.4 \\

   &
  SOLD2 &
  Line &
  \textbf{5.9} &
  29.6 &
  64.2 &
  76.6 &
  79.6 &
  \textbf{88.1} &
  \textbf{96.4} &
  \textbf{97.9} &
  72.9 &
  84.0 &
  86.7 \\ \hline
  \multirow{3}{*}{NEF~\cite{Ye_2023_CVPR}} &
  PiDiNet$\dagger$ &
  Bézier &
  11.9 &
  16.9 &
  11.4 &
  62.0 &
  91.3 &
  15.7 &
  68.5 &
  96.3 &
  13.0 &
  64.6 &
  93.3 \\
   &
  PiDiNet &
  Bézier &
  15.1 &
  16.5 &
  11.7 &
  53.3 &
  89.8 &
  13.6 &
  52.2 &
  89.1 &
  12.3 &
  51.8 &
  88.7 \\
   &
  DexiNed &
  Bézier &
  21.9 &
  15.7 &
  11.3 &
  48.3 &
  87.7 &
  11.5 &
  39.8 &
  71.7 &
  10.8 &
  42.1 &
  76.8 \\ \cline{1-14} 
  \multirow{2}{*}{EMAP~\cite{Li2024CVPR}} &
  PiDiNet &
  Bézier / Line &
  9.2 &
  15.6 &
  30.2 &
  75.7 &
  89.8 &
  35.6 &
  79.1 &
  95.4 &
  32.4 &
  77.0 &
  92.2 \\
  &
  DexiNed &
  Bézier / Line &
  8.8 &
  8.9 &
  56.4 &
  88.9 &
  94.8 &
  62.9 &
  89.9 &
  95.7 &
  59.1 &
  88.9 &
  94.9\\
\cline{1-14} 
\multirow{2}{*}{EdgeGaussians~\cite{chelani2024edgegaussians}} &
  PiDiNet &
  Bézier / Line &
  11.7 &
  10.3 &
  17.1 &
  73.9 &
  83.1 &
  26.0 &
  87.2 &
  92.5 &
  20.6 &
  79.3 &
  86.7 \\
  &
  DexiNed &
  Bézier / Line &
  9.6 &
  8.4 &
  42.4 &
  91.7 &
  95.8 &
  49.1 &
  94.8 &
  96.3 &
  45.2 &
  93.7 &
  95.7 \\
  \cline{1-14} 
\multirow{2}{*}{Ours} &
  PiDiNet &
  Bézier / Line &
  10.8 &
  12.3 &
  33.3 &
  83.9 &
  92.7 &
  30.7 &
  78.8 &
  95.5 &
  31.5 &
  80.8 &
  93.7 \\
  &
  DexiNed &
  Bézier / Line &
  8.2 &
  \textbf{7.5} &
  \textbf{69.6} &
 \textbf{ 93.8 }&
  \textbf{96.3} &
  79.0 &
  94.9 &
  96.7 &
 \textbf{ 73.7 }&
 \textbf{ 94.0} &
 \textbf{ 96.2}\\
  \hline
\end{tabular}
}
\vspace{-0.5em}
\caption{\textbf{Quantitative results on ABC-NEF~\cite{Ye_2023_CVPR}.} Results from NEF's released pretrained models are marked with $\dagger$.  Our method outperforms all recent learning-based 3D line/curve reconstruction approaches across all metrics, while achieving precision comparable to the line-mapping system, LIMAP.
}

\label{tab:ABC}
\vspace{-8px}
\end{table*}

\noindent \textbf{Metrics.}
Following metrics established in~\cite{chelani2024edgegaussians,Li2024CVPR}, we assess performance using Accuracy (Acc.), Completeness (Comp.) measured in millimeters (mm), as well as Recall ($R_\tau$ ), Precision ($P_\tau$), F-score ($F_\tau$) expressed in percentages, with a distance threshold $\tau$ in mm. The metrics are computed by sampling points in
proportion to the parameterized edge lengths and downsampling them at the same resolution as the ground-truth edges. Additionally, to compare the efficiency, we provide training times, and to demonstrate the compactness of the proposed parameterized curves, we report the number of curves/lines generated.

\noindent \textbf{Implementation Details.} 
Our method is trained for 10k iterations for each test instance. To ensure meaningful updates, we delay the introduction of adaptive control strategies for parametric edges until later stages of training, as these edges require sufficient optimization to converge to a suitable form. Specifically, curve linearization is initiated at the 3k-th iteration. From the 7k-th iteration onward, we apply curve linearization and merging strategies to refine the parametric curves. To balance efficiency, operations of splitting, linearization, and merging are performed every 1k iterations. Additionally, we optimize the curve opacity before the 7k-th iteration. After that, the opacity attribute is fixed, and the mask loss is applied.

\subsection{Evaluations}

\begin{table}[!t]
\centering
\scalebox{0.8}{
\begin{tabular}{c|c|ccc}
\hline
   Method & Runtime $\downarrow$      & $N_{\text{Bézier}} $ $\downarrow$ &$N_{\text{line}}$$\downarrow$  & $N_{\text{sum}}$$\downarrow$  \\
\hline
NEF$\dagger$~\cite{Ye_2023_CVPR}                 & 1.5 hours &22.9 & 0 &22.9\\

EMAP~\cite{Li2024CVPR}                  &  2.5 hours & 9.2 & 36.6 & 45.8\\
EdgeGaussians ~\cite{chelani2024edgegaussians}       &  4 mins &13.0&84.9 & 97.9 \\
Ours                        & 3 mins   & 6.9 &  22.0   & 28.9 \\
\hline
\end{tabular}

}
\vspace{-0.5em}
\caption{\textbf{Runtime and number of curves/lines comparisons on ABC-NEF~\cite{Ye_2023_CVPR}.}  Results from NEF's released pretrained models are indicated by $\dagger$. $N_{\text{Bézier}}$, $N_{\text{line}}$ and $N_{\text{curve}}$ represent the average number of reconstructed Bézier, lines, and total curve edges.
}

\label{tab:ABC_2}
\vspace{-1.em}
\end{table}

\begin{figure*}[tbh]
  \centering
  \includegraphics[width=1.0\textwidth]{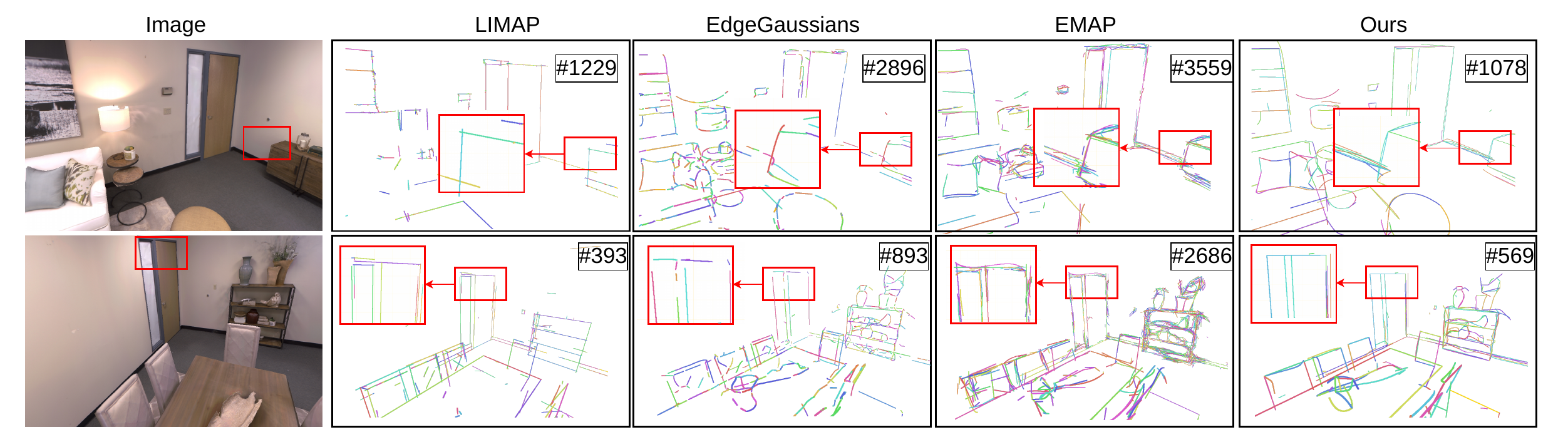}
    \vspace{-16px}
  \caption{\textbf{Qualitative comparisons on Replica~\cite{replica19arxiv}.} Distinct colors represent different parametric edges, with the numbers in the images indicating the count of parametric edges. Our method achieves accurate edge reconstruction,  balancing compactness and completeness. }
\vspace{-6px}
  \label{fig:replica_vis}
\end{figure*}

\begin{figure*}
  \centering
  \includegraphics[width=0.98\textwidth]{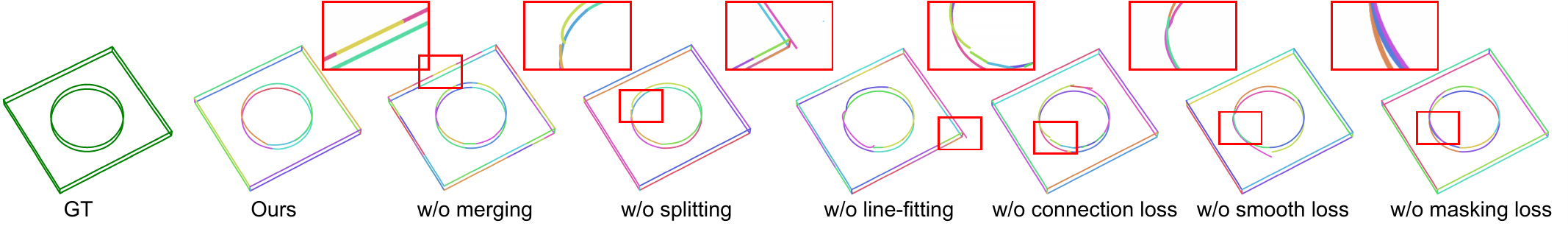}
  \caption{\textbf{Qualitative ablations on key components of our method.}  Excluding critical designs from the full version introduces additional noise, incompleteness, or redundancy in the qualitative reconstruction results.}
  \label{fig:bla-vis}
  \vspace{-8px}
\end{figure*}

\textbf{Evaluation on ABC-NEF Dataset~\cite{Ye_2023_CVPR}.} The qualitative and quantitative comparisons are presented in Fig.~\ref{fig:abc_vis} and Tab.~\ref{tab:ABC}, respectively. Following EMAP~\cite{Li2024CVPR}, we quantitatively report the results using PiDiNet~\cite{su2021pdc} and DexiNed~\cite{poma2020dense} as edge detectors, while the visualized results demonstrate the performance of DexiNed. LIMAP~\cite{liu2023limap} achieves impressive precision but lacks the ability to represent curves, often resulting in significant edge omissions. NEF~\cite{Ye_2023_CVPR} exhibits lower accuracy due to limitations in point sampling and fitting-based post-processing. EMAP struggles with incomplete reconstructions, resulting in lower scores in completeness and recall. EdgeGaussians~\cite{chelani2024edgegaussians} frequently produces fragmented reconstructions, due to inaccuracies in estimating principal axes of constructed Gaussians, which cause the subsequent fitting to converge to suboptimal local minima. 
Our method directly optimizes parametric edge representations, inherently incorporating local edge constraints and eliminating errors introduced by fitting stages. As a result, the proposed method
outperforms all learning-based methods 
across all metrics, while achieving precision on par with LIMAP. 

To further compare the efficiency and compactness of parametric edge extraction, we also evaluate runtime and the number of parametric edges using DexiNed as the image edge detector. Both EdgeGaussians and our method significantly outperform implicit-representation-based approaches like NEF and EMAP in terms of computational efficiency. NEF utilizes the fewest edges due to its frequent edge omissions, as reflected in its low recall scores. 
Our method achieves compact parametric edge representation with efficient reconstruction while delivering complete and highly accurate edge reconstruction. It shows a superior balance of efficiency, compactness, and accuracy.

\noindent \textbf{Evaluation on MV2Cyl's real objects ~\cite{hong2024mv2cyl}.} 
Real-world datasets often pose significant challenges to algorithms due to inconsistencies in edge detection across different viewpoints and non-uniform coverage of objects. Since NEF and EMAP fail to produce reasonable 3D edges in our experiments, we compare our approach with EdgeGaussian. As shown in Fig.~\ref{fig:teaser}, our method demonstrates better reverse engineering capabilities. EdgeGaussian generates a significant amount of noisy Gaussian distributions due to the introduction of 2D edge noise caused by variations in lighting and viewpoint in real-world scenes. In contrast, our method inherently enforces local consistency constraints and better mitigates the impact of noise. We follow MV2Cyl~\cite{hong2024mv2cyl} and use ICP to align the reconstructed edge point cloud with the ground-truth CAD model. Tab.~\ref{tab:real_demo} shows that our method achieves higher accuracy and more compact 3D edges.

\begin{table}
\centering
\scalebox{0.70}{
\begin{tabular}{c|cc|ccc|c}
\hline
   Method & Acc. $\downarrow$      & Comp. $\downarrow$ & $\text{R}_{10} \uparrow$  &  $\text{P}_{10} \uparrow$  &  $\text{F}_{10} \uparrow$  &$ N_{\text{curve}}$ $\downarrow$  \\
\hline

EdgeGaussians ~\cite{chelani2024edgegaussians}  &  13.8 &\textbf{7.7}& 74.4 & 46.1& 56.7 & 1952.5 \\
Ours  & \textbf{8.5}& 7.8& \textbf{74.6}&\textbf{ 71.6}& \textbf{73.1}& \textbf{70.5} \\ 
\hline
\end{tabular}

}
\vspace{-0.5em}
\caption{\textbf{Quantitative comparisons on  MV2Cyl's real objects.} 
}

\label{tab:real_demo}
\vspace{-10px}
\end{table}

\begin{table}
\centering
\scalebox{0.63}{

\begin{tabular}{l|cccccccc}
\hline
 Method & Acc.$\downarrow$ & \multicolumn{1}{l}{Comp.$\downarrow$} &  $\text{R}_5 \uparrow$ &  $\text{P}_5\uparrow$  &  $\text{F}_5 \uparrow$ & $N_\text{curve}$$\downarrow$ & $N_\text{line}$$\downarrow$ & $N_\text{sum}$$\downarrow$  \\ \hline
 Ours                & 8.2 & 7.5          & 69.6    & 79.0   & 73.7 &6.9 & 22.0 & 28.9 \\ \hline

 w/o merging   & 7.9  & 7.5 & 70.4 & 79.6 & 74.4 & 8.4&63.3   &71.7    \\
 w/o splitting & 8.3 & 7.5 & 69.6 & 78.8 & 73.4  &6.7& 20.5&27.2\\   

 w/o line-fitting & 8.5 & 7.4 & 63.2 & 75.1 & 68.4&0&32&32   \\  

 w/o connection loss   & 8.2         &  7.4         &  69.7    & 78.9    &  73.2& 5.5&26.3 &31.8         \\
 w/o smooth loss   &  8.4 & 7.5 & 69.7 &78.5& 73.5 & 7.1&24.3&31.4       \\
 w/o masking loss   &8.9  & 7.1 & 74.7 & 78.2 & 76.1& 10.1 &25.3   &35.4   \\

\hline
\end{tabular}
}
\vspace{-5px}
\caption{\textbf{Ablations on parametric edge extraction components on ABC-NEF~\cite{Ye_2023_CVPR} with DexiNed~\cite{poma2020dense} edge map detection.} }
\label{tab:abl}
\vspace{-10px}
\end{table}
\noindent \textbf{Evaluation on Replica Dataset~\cite{replica19arxiv}.}
To demonstrate the capability of our method in capturing scene-level geometry, we further evaluate it on the Replica dataset. Given that NEF struggles to handle large-scale scenes effectively, we exclude its reconstruction results from this experiment. Due to the absence of ground-truth 3D edges, we focus solely on qualitative comparisons to evaluate the performance of different methods.  As shown in Fig.~\ref{fig:replica_vis}, we compare the visualized edges and the number of reconstructed edges across different methods. Our method maintains its superiority even at the scene level, benefiting from the direct optimization of parametric curve representations.  Our method produces a similar number of edges to LIMAP, which suffers from severe edge omissions. Compared to EdgeGaussians and EMAP, our approach achieves comparable reconstruction accuracy while using fewer parametric edges, ensuring a compact and efficient representation. This facilitates downstream applications and enables a more interpretable understanding of the reconstructed edges.

\subsection{Ablations and Analysis}
We conduct ablations to validate the effectiveness of key loss functions and adaptive parametric curve control strategies, as presented in Tab.~\ref{tab:abl} and Fig.~\ref{fig:bla-vis}. Ablations are conducted on the ABC-NEF benchmark, with DexiNed as the edge detector. 
First, removing the curve merge operation leads to a slight increase in recall and precision but results in a significant rise in the number of reconstructed curves, which is undesirable for achieving compact representations. Second, disabling curve splitting introduces noisy, jagged curves, demonstrating its necessity for noise control. Third, eliminating the straight-line fitting operation causes a noticeable decline in both accuracy and completeness, as straight lines are prevalent in man-made objects and are easier to merge, contributing to overall simplicity. Fourth, removing the loss function that encourages curve endpoint connections reduces the reconstruction accuracy and results in less compact parametric curve representations. Fifth, smoothness loss is critical for improving reconstruction accuracy and reducing noise. Lastly, the masking loss effectively eliminates redundant curves. Without this loss, the method tends to produce overlapping curves and introduces more noisy edges. Integrating all components, our parametric edge extraction pipeline achieves an optimal balance between accuracy, completeness, and compactness.

\section{Conclusion}
This paper presents a one-stage 3D parametric curve reconstruction method based on a bi-directional coupling mechanism between parametric curves and edge-oriented Gaussians. The limitation is that the method remains contingent on the quality of 2D edges, and it may be mitigated through 2D foundation models for estimating better edge maps. Our method also provides a potential pathway and future direction for lifting 2D edge detection foundation models into 3D, to benefit 3D visual and geometric reasoning.

\section*{Acknowledgements}
This work is supported in part by the NSFC (62325211, 62132021, 62372457), the Major Program of Xiangjiang Laboratory (23XJ01009), Young Elite Scientists Sponsorship Program by CAST (2023QNRC001), the Natural Science Foundation of Hunan Province of China (2022RC1104).
{
    \small
    \bibliographystyle{ieeenat_fullname}
    \bibliography{main}
}
\maketitlesupplementary

This supplementary material first provides detailed experimental settings, including data processing procedures, implementation details for comparison baseline methods, and evaluation metrics. 

\begin{figure*}[tbh]
  \centering
  \includegraphics[width=0.95\textwidth]{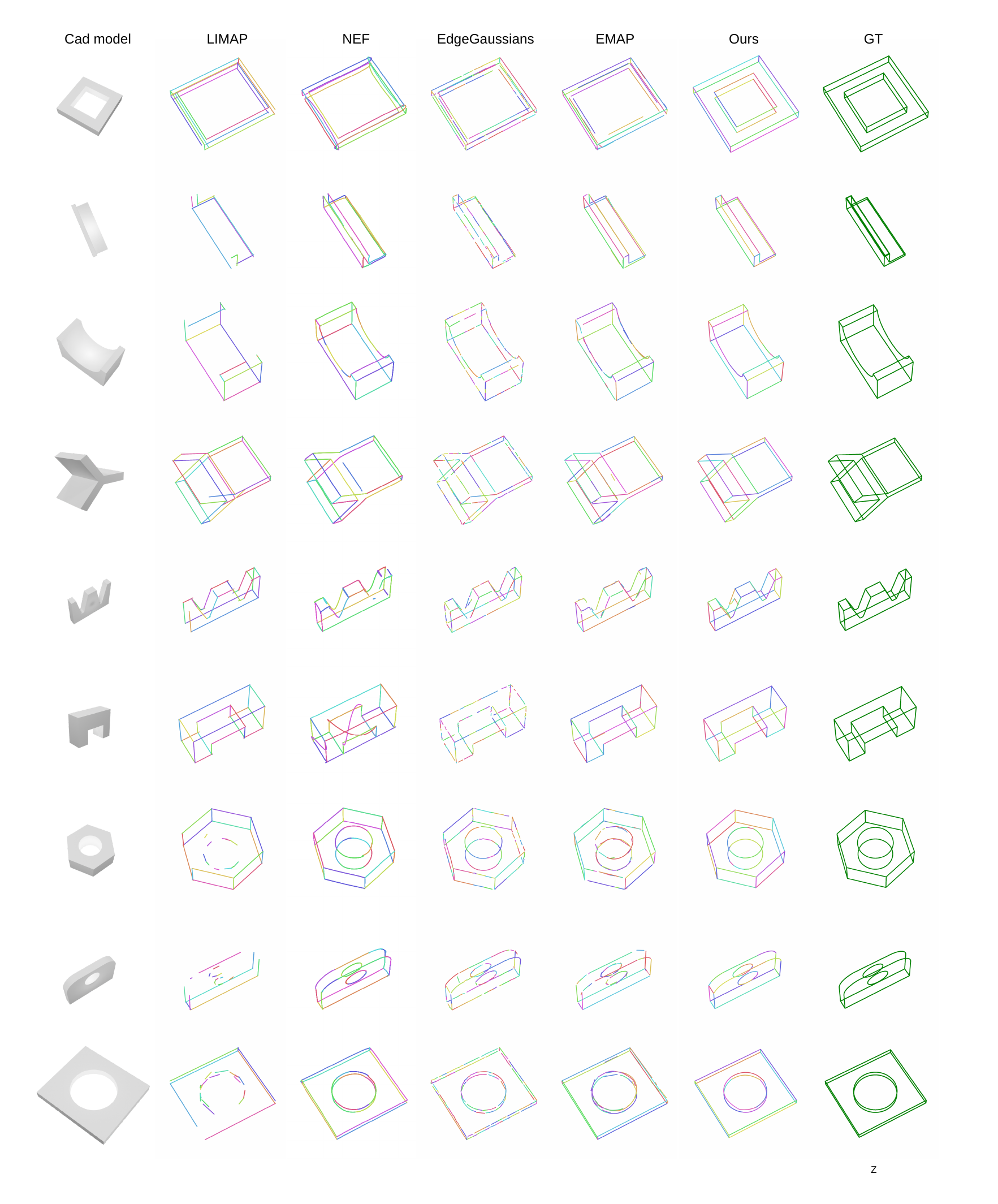}
  \caption{\textbf{Qualitative comparisons on ABC-NEF~\cite{Ye_2023_CVPR}.} Distinct colors represent different curves. Our method achieves more complete and accurate edge reconstruction of objects while maintaining parametric compactness.}
  \label{fig:supp_abc_vis}
\end{figure*}

In Fig.~\ref{fig:supp_abc_vis}, we provide additional visual comparisons of our method against state-of-the-art baselines on ABC-NEF~\cite{Ye_2023_CVPR}. Our code and data are available at \url{https://github.com/zhirui-gao/Curve-Gaussian}.

\label{sec:exp_setting}
\section{Datasets}
The proposed method is evaluated on three publicly available datasets: ABC-NEF~\cite{Ye_2023_CVPR}, Mv2Cyl's Real Objects~\cite{hong2024mv2cyl}, and the Replica Dataset~\cite{replica19arxiv}. Detailed descriptions of each dataset and the experimental setups are provided below.

\textbf{{ABC-NEF Dataset.}} The ABC-NEF dataset is a widely adopted benchmark for evaluating 3D curve reconstruction quality. It contains precise CAD models with diverse curve types, comprising 115 objects in total. Following the protocols of EMAP~\cite{Li2024CVPR} and EdgeGaussians~\cite{chelani2024edgegaussians}, objects with indistinct sharp curve features were filtered out, resulting in 82 objects for evaluation. Each object includes 50 images with a resolution of $800 \times 800$. For computational efficiency, all images were resized to $400 \times 400$.

\textbf{Mv2Cyl's Real Objects Dataset.} The Mv2Cyl's Real Objects dataset consists of multiple 3D-printed objects captured using an iPhone 12. Multi-view images were extracted from continuous videos, with camera poses computed using COLMAP. Ground truth CAD models are also provided for evaluation.  This dataset presents a significant challenge for 3D edge detection due to the top-down perspective of the captured images. To generate edge maps, the following pipeline is employed:
\begin{itemize}
    \item SAM2~\cite{ravi2024sam2} is utilized to segment objects from the background.
    \item A monocular normal estimation network~\cite{ye2024stablenormal} is applied to identify high-curvature regions as object edges. This approach demonstrated superior performance compared to edge detection methods~\cite{su2021pdc,poma2020dense}, which often extracted irrelevant edges due to lighting interference.
    \item Edge maps are resized to $480 \times 480$ for faster processing.
\end{itemize}

Since the camera poses and CAD models are not aligned in the same coordinate system, the Iterative Closest Point (ICP) was used to register the reconstructed edge points with the ground truth CAD curves for quantitative evaluation. 

The lack of real-world benchmarks for 3D curve reconstruction is a notable gap in the field. The Mv2Cyl's Real Objects dataset addresses this limitation.  After getting permission from the authors of Mv2Cyl, a standardized benchmark for evaluating 3D edge reconstruction will be proposed based on their multi-view images. We believe it will be an important contribution to the field.

\textbf{Replica Dataset.}
We follow the experimental setup in EMAP~\cite{Li2024CVPR} on this dataset, focusing on three scenes: Room 0, Room 1, and Room 2.

\section{Baselines} 

Our method is compared against four state-of-the-art 3D line and curve reconstruction baselines. These include three learning-based methods—NEF~\cite{Ye_2023_CVPR}, EdgeGaussians~\cite{chelani2024edgegaussians}, and EMAP~\cite{Li2024CVPR}—and one line-based Structure-from-Motion (SfM) method, LIMAP~\cite{liu2023limap}. For a fair comparison, the default parameters and settings provided by the authors are adopted for all baselines. In the case of NEF and EMAP, their pre-trained models were directly applied to generate visual results.

\section{Evaluation Metrics} 
To quantitatively assess the performance of our method, we adopt a set of evaluation metrics that align with established protocols in this field. Points are uniformly sampled along both the reconstructed parametric curves and the corresponding ground-truth edges, enabling a direct comparison between them. The metrics are introduced as follows:

\begin{itemize}
    \item \textbf{Accuracy}: This metric calculates the average distance from each predicted point to its closest counterpart on the ground-truth curve. Smaller values correspond to higher accuracy.
    \item \textbf{Completeness}: This measures the average distance from each ground-truth point to the nearest predicted point. Improved performance is indicated by lower values.
    \item \textbf{Precision at Threshold $\tau$ ($P(\tau)$)}: This quantifies the proportion of predicted points that lie within a distance $\tau$ of any ground-truth point. Higher precision values reflect better alignment with the ground truth.
    \item \textbf{Recall at Threshold $\tau$ ($R(\tau)$)}: This evaluates the proportion of ground-truth points that have at least one predicted point within a distance $\tau$. Higher recall values signify better coverage of the ground truth.
\end{itemize}

In addition to these conventional metrics, we introduce a new metric, \textbf{Curve Count}, which evaluates the compactness of the reconstructed curves, which can be found in Table 2 of the main paper. This metric counts the total number of curves generated by the method, providing insight into the efficiency of the representation. A smaller curve count indicates a more concise and compact reconstruction, which is advantageous for downstream tasks that require efficient curve processing.

For consistency with prior evaluations, precision and recall are computed at distance thresholds of $\tau = 5$, $10$, and $20$ millimeters (mm). For other parameters in the evaluation, such as the number of sampling points, we adhere to the same settings as those used in EMAP~\cite{Li2024CVPR}.

\section{Implementation Details} 
\label{sec:imp_detail}
The weight coefficients $\lambda_1$, $\lambda_2$, $\lambda_3$, and $\lambda_4$ were set to $0.01$, $0.01$, $0.01$, and $0.0005$, respectively. For all Bézier curves and straight lines, a default of 12 Gaussian points was sampled per curve. 

For the ABC dataset, the midpoints of Bézier curves were initialized by uniformly sampling $15 \times 15 \times 15$ points in 3D space. In contrast, for COLMAP-based datasets, the midpoints were initialized using the point cloud generated by Structure-from-Motion (SfM). The threshold for curve merging was kept consistent with EMAP~\cite{Li2024CVPR}. 

To ensure the quality of the reconstructed curves, curves with an opacity below $0.05$ were removed, and curves with bending angles exceeding $20^\circ$ were split. Additionally, Gaussian components with mask attributes below $0.01$ were considered redundant and discarded. Further details on parameter design and implementation can be found in the accompanying code.

\end{document}